\documentclass{article}
\usepackage{spconf,amsmath,amssymb,epsfig,booktabs,multirow,hyperref}

\let\OLDthebibliography\thebibliography
\renewcommand\thebibliography[1]{
  \OLDthebibliography{#1}
  \setlength{\parskip}{0pt}
  \setlength{\itemsep}{0pt plus 0.3ex}
}

\begin{document}\sloppy

% Example definitions.
% --------------------
\def\x{{\mathbf x}}
\def\L{{\cal L}}

% Title.
% ------
\title{Enhancing Lip Reading with Multi-Scale Video and Multi-Encoder}

% Single address.
% ---------------
\name{He Wang$^1$, Pengcheng Guo$^1$, Xucheng Wan$^2$, Huan Zhou$^2$, Lei Xie$^{1{\ast}}$ \thanks{*Corresponding author}}
%Address and e-mail should NOT be added in the submission paper. They should be present only in the camera ready paper. 
\address{$^1$Audio, Speech and Language Processing Group (ASLP@NPU), School of Computer Science,\\Northwestern Polytechnical University, Xian, China\\$^2$IT Innovation and Research Center, Huawei Technologies}

\maketitle

\begin{abstract}
Automatic lip-reading (ALR) aims to automatically transcribe spoken content from a speaker's silent lip motion captured in video. 
% Current mainstream ALR systems still use Transformers or Conformers as encoders for visual feature modeling. 
% Moreover, research on the size of the lip region of interest (ROI) remains scarce. 
Current mainstream lip-reading approaches only use a single visual encoder to model input videos of a single scale.
In this paper, we propose to enhance lip-reading by incorporating multi-scale video data and multi-encoder. 
% Specifically, we design different ROI scales according to the speaker's face size for extracting lip motion video. 
% On multi-system building and fusion, we propose the Enhanced ResNet3D visual front-end (VFE) module for lip feature extraction, introduce the recently proposed Branchformer and E-Branchformer as visual encoders, and employ the recognizer output voting error reduction (ROVER) to fuse texts transcribed from all ALR systems. 
Specifically, we first propose a novel multi-scale lip motion extraction algorithm based on the size of the speaker's face and an Enhanced ResNet3D visual front-end (VFE) to extract lip features at different scales.
For the multi-encoder, in addition to the mainstream Transformer and Conformer, we also incorporate the recently proposed Branchformer and E-Branchformer as visual encoders.
In the experiments, we explore the influence of different video data scales and encoders on ALR system performance and fuse the texts transcribed by all ALR systems using recognizer output voting error reduction (ROVER).
Finally, our proposed approach placed second in the ICME 2024 ChatCLR Challenge Task 2, with a 21.52\% reduction in character error rate (CER) compared to the official baseline on the evaluation set.
\end{abstract}

\section{introduction}
\label{intro}
\begin{figure*}[htbp]
  \centering
  \centerline{\includegraphics[width=16cm]{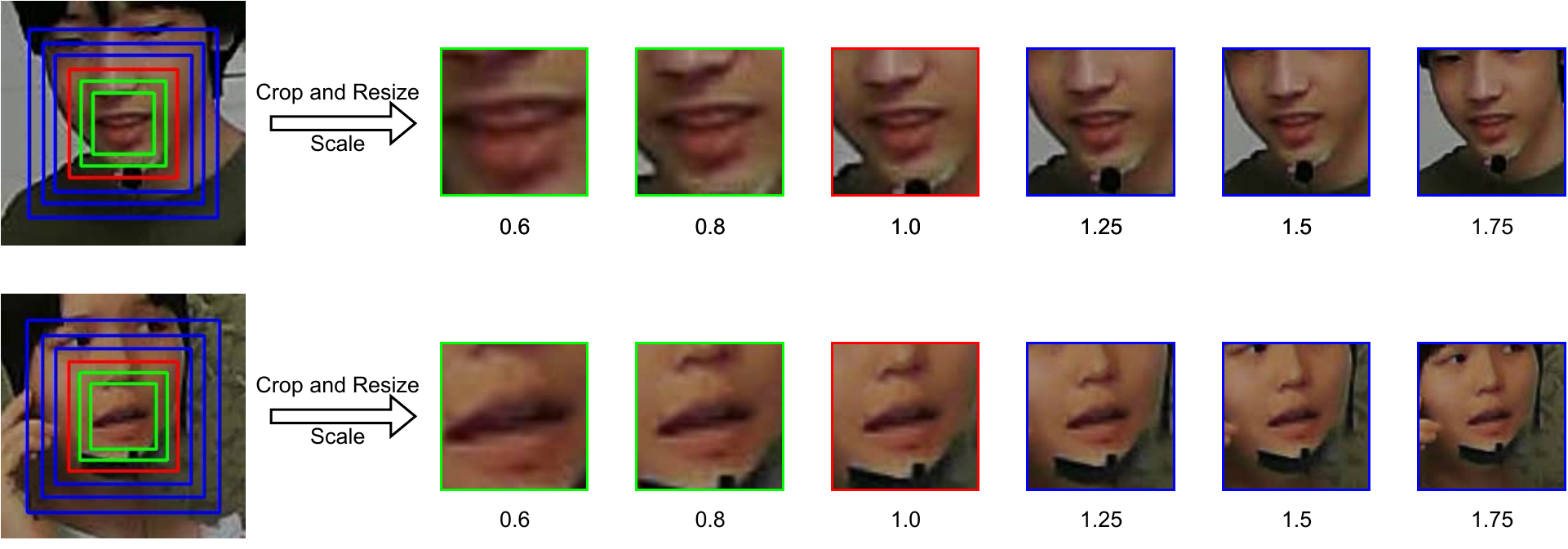}}
\caption{Examples of multi-scale lip motion videos of speaker S217 (top) and S443 (bottom) from the ChatCLR training set.}
\label{fig:1}
\end{figure*}
With the development of deep learning techniques, automatic speech recognition (ASR) has progressed significantly, already reaching or even surpassing the human level on some open-source benchmarks~\cite{xiong2016achieving}. 
% However, the effectiveness of ASR systems heavily relies on the quality of the input audio signal.
However, the performance of ASR systems tends to degrade noticeably in real-world scenarios with noisy backgrounds or distant recording conditions, such as family gatherings, cocktail parties, or multi-person meetings.
Moreover, in applications like silent speech interfaces or assisting individuals with speech disorders, automatic lip-reading (ALR) has attracted increasing attention.

Many researches have been conducted on ALR or audio-visual speech recognition (AVSR), which integrates visual features into the ASR system.
In the early stage, Transformer~\cite{vaswani2017attention}, with its powerful sequence modeling capabilities, is widely used as the backbone for visual modeling in many studies~\cite{lee2020audio,serdyuk2021audio,serdyuk2022transformer}. 
However, as the Conformer~\cite{gulati2020conformer} proposed, it has gradually replaced Transformer and become the mainstream choice for visual backbone, no matter in ALR~\cite{liu2022end,chang2024conformer} or AVSR~\cite{ma2021end,ma2023auto} studies.
Despite the surge in research on ALR or AVSR in recent years, most studies have focused on English, with less research dedicated to Chinese lip-reading.
The absence of a large dataset for Chinese lip-reading was undoubtedly a significant reason, leaving researchers without a benchmark for validation and fair comparison. 
Since 2021, the Multimodal Information based Speech Processing (MISP) Challenge series~\cite{chen2022first,wang2023multimodal,wu2024multimodal} has been successfully held for three consecutive editions to advance the research on Chinese audio-visual speech processing and released a large and challenging Chinese audio-visual dataset~\cite{chen2022first,chen2022audio}.
% The scope of tasks includes key speech technologies, such as audio-visual keyword detection, speaker diarization, speech recognition, and target speaker extraction. 
In addition, the hosting of the Chinese Continuous Visual Speech Recognition Challenge (CNVSRC) 2023~\footnote{http://cnceleb.org/competition} further promotes research in Chinese lip-reading, releasing a larger volume (over 300 hours) of Chinese audio-visual dataset~\cite{chen2023cn}.
However, it mainly focuses on professional announcers, speechmakers, or bloggers with prepared topics, limiting some practical applicability.

To promote universal Chinese lip-reading in real-world scenarios, the ICME 2024 Chat-scenario Chinese Lipreading (ChatCLR) Challenge~\footnote{https://mispchallenge.github.io/ICME2024} is launched. 
The target speaker lip-reading track (Task 2) is designed for lip-reading in the free talk scenarios, and over 100 hours of far-field real-recorded video data are released.
This paper elaborates on our proposed lip-reading approach that builds and fuses multiple ALR systems with different visual encoders and multi-scale lip motion video data.
For data extraction, we design 6 different lip regions of interest (ROIs) according to the size of the speaker's face and the coordinate of the lip. 
Based on our previous works~\cite{wang2024mlca,wang2024npu}, we propose an Enhanced ResNet3D visual front-end (VFE) module for better visual feature extraction. 
Moreover, recently proposed Branchformer~\cite{peng2022branchformer} and E-Branchformer~\cite{kim2023branchformer} encoders are introduced as the backbone for visual modeling, constructing more diverse ALR systems with different scales of video data.
Finally, the recognizer output voting error reduction~\cite{fiscus1997post} (ROVER) technique is used to fuse the transcripts from all ALR systems. 
Overall, our contributions are as follows:
\begin{itemize}
    % \item We design different lip ROIs based on the speaker's face size and explore the influence of lip motion video scale on ALR system performance.
    \item We introduce an algorithm to extract multi-scale lip motion videos based on the size of the speaker's face.
    \item We propose an Enhanced ResNet3D visual front-end module based on the ResNet~\cite{he2016deep} and 3D convolutions.
    % \item We systematically compare the recently proposed Branchformer and E-Branchformer with mainstream Transformer and Conformer to explore the most effective encoder for far-field Chinese ALR system.
    \item We systematically compare different visual encoders and video data scales on the performance of ALR.
    \item Our proposed approach achieves a character error rate (CER) of 78.17\% on the ChatCLR Challenge Task 2 evaluation set, yielding a reduction of 21.52\% CER compared to the official baseline, ranking second place.
\end{itemize}

\section{Method}
\subsection{Multi-Scale Lip Video Data Extraction}
Many researches~\cite{zhang2020can,wang2024npu} have indicated the performance of lip-reading recognition systems is not solely determined by the motion of the speaker's lips. 
Visual information from the cheeks, chin, and even forehead can improve recognition accuracy. 
Moreover, the winning approach proposed by Wang et al.~\cite{wang2024npu} in CNVSRC 2023 also shows the performance of ALR systems improves with an increase in the field of view of the video data centered around the lips. 
However, their method of enlarging the lip ROI by a fixed adjustment of the crop size for each video frame (from 48 to 112) does not consider the discrepancy among all speakers' face sizes or the distances between the camera and speakers. 
This could lead to inconsistencies in the facial area covered by the obtained lip motion videos, even when using the same crop size. 

Therefore, we design a lip motion extraction method based on the size of the speaker's face in the video. 
The official ChatCLR dataset provides all speakers' face and lip coordinates on the image in almost every video frame.
For a video sequence of speaker $\text{S}$ with a total number of frames $T$ (25 frames per second), we define $t$ as the number of frames where the speaker's face and lip are successfully detected ($t \le T$). 
The coordinates of the top-left and bottom-right corners of the $i$-th detected face are defined as $(Lface^i_x, Lface^i_y)$ and $(Rface^i_x, Rface^i_y)$, respectively. 
Thus, the crop size $L$, that is, the side of the square lip ROI, can be formulated as:
\begin{equation}
    \begin{gathered}
        W^i = Rface^i_x - Lface^i_x \\
        H^i = Rface^i_y - Lface^i_y \\
        L = \frac{1}{t}\sum_{i=1}^{t}{\frac{(W^i + H^i)}{8}} \times scale
    \end{gathered}
\end{equation}
where $scale$ is a scaling factor; in this paper, we employ 6 factors: 0.6, 0.8, 1.0, 1.25, 1.5, and 1.75.

Since we adopt a lip-centered strategy for extracting lip motion videos, the lip ROI can be determined once the lip center coordinates for each frame are identified. 
We define the top-left and bottom-right coordinates of the detected lip of speaker $\text{S}$ in the $i$-th frame as $(Llip^i_x, Llip^i_y)$ and $(Rlip^i_x, Rlip^i_y)$, respectively. 
Then, the formula to calculate the lip center coordinates (LCC) for each frame becomes:
\begin{equation}
LCC = \left(\frac{Llip^i_x + Rlip^i_x}{2}, \frac{Llip^i_y + Rlip^i_y}{2}\right)
\end{equation}
In cases where the speaker's face is obscured or not directed towards the camera, it becomes hard to detect their lips. 
To overcome this, for a failed lip detection frame, we search for the nearest detected lip forward or backward in time.
This allows us to calculate the center coordinates for every frame and extract lip motion video data as accurately as possible. 
It is worth noting that if the detection rate of the face or lip in a segment of the video does not exceed 50\%, it will be discarded. 
All lip motion data extracted at any scale will be resized to a uniform size of $112 \times 112$.

Figure~\ref{fig:1} shows the lip motion video extracted from two speakers under different scaling factors. 
It can be observed that when the $scale$ is set to 0.6 or 0.8, the extracted lip motion data almost only contains the speaker's lips. 
As the value of $scale$ increases, areas such as the chin, cheeks, nose, and eyes gradually enter the lip-centered ROI.

\subsection{Enhanced ResNet3D Visual Front-end}
\begin{figure}[t]
  \centering
  \centerline{\includegraphics[width=8.5cm]{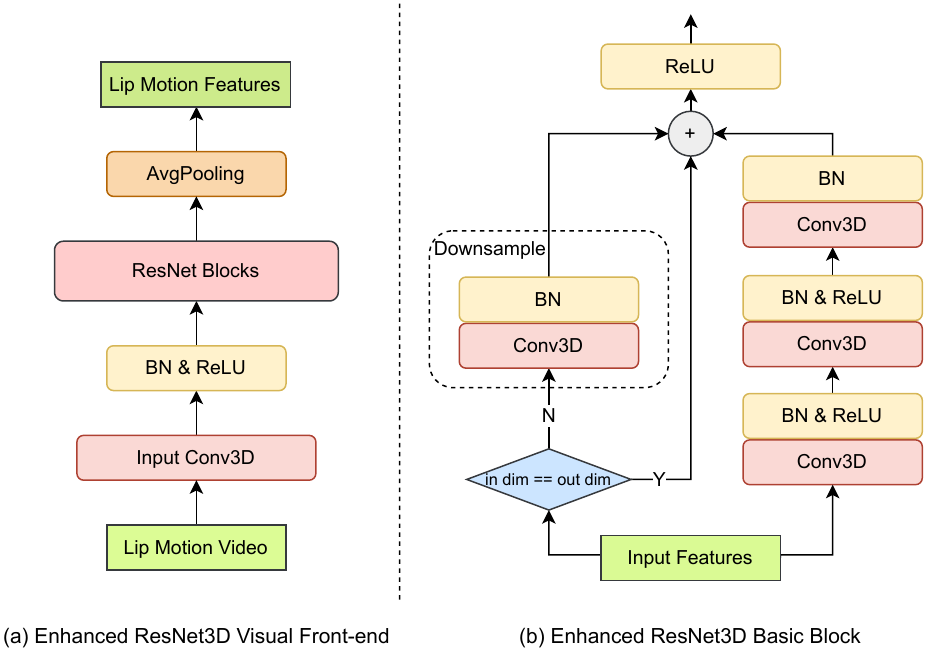}}
\caption{Detailed structures of the proposed Enhanced ResNet3D visual front-end (a) and its basic block (b).}
\label{fig:2}
\end{figure}
Figure~\ref{fig:2} shows the detailed structures of our proposed Enhanced ResNet3D visual front-end.
The overall module design is inspired by the classic image feature extraction network ResNet~\cite{he2016deep}, with the difference being that we replace the 2-dimensional convolution in ResNet with 3-dimensional convolution (Conv3D) to model the input 3D video data. 
The Enhanced visual front-end mainly consists of three parts (as shown in Figure 2a), the input Conv3D for mapping the feature channels of the input video data to a higher dimension, AvgPooling for averaging the video height and width dimensions at the final, and the ResNet3D blocks in the middle part modeling the visual features.
Each layer of the ResNet3D block comprises several basic blocks, as detailed in Figure 2b. 
These blocks are primarily comprised of stacked Conv3D and batch norm, forming the video feature modeling unit. 
Additionally, the first basic block in each ResNet3D block performs a two-fold down-sampling of the input visual features in height and width and maps to higher feature channels. 
Therefore, the down-sample module in the basic block only takes effect for the first basic block in each ResNet3D block.

\subsection{Multi-System Building and Fusion}
\begin{figure}[t]
  \centering
  \centerline{\includegraphics[width=8.5cm]{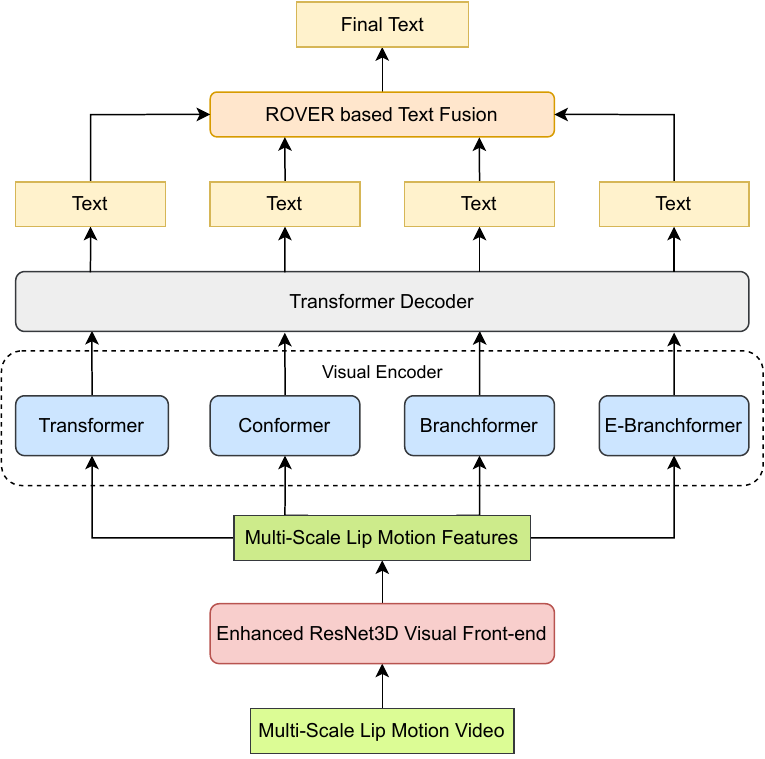}}
\caption{Block diagram of the proposed multi-system fusion approach for automatic lip-reading.}
\label{fig:3}
\end{figure}
Figure~\ref{fig:3} shows our proposed lip-reading approach based on multi-scale lip motion video data and multi-system fusion. 
On one hand, to build ALR systems as diverse as possible, we fix the model structure by using the same visual encoder and feeding different scales of video to build ALR systems.
On the other hand, we use video data of one scale and vary the visual encoder.
In this paper, both strategies are employed, and both achieve commendable improvements (refer to Section \ref{multi-system-fusion-improvement}). 
Each ALR system also shares some commons: they all use our proposed Enhanced ResNet3D visual front-end module for extracting features from raw lip motion video, and they all employ a Transformer-based decoder for the transcription.
Additionally, inspired by the ASR system based on joint connectionist temporal classification (CTC) and attention proposed by Kim et al.~\cite{kim2017joint}, all ALR systems also adopt a similar scheme, with the model loss defined as follows:
\begin{equation}
    loss= \lambda loss_{ctc} + (1-\lambda) loss_{attention},
\end{equation}
where $loss_{ctc}$ is the CTC loss computed from the encoder's output, used for learning the temporal alignment between the encoder's output and text label; $loss_{attention}$ is the cross-entropy (CE) loss calculated from the decoder's output and text labels.
$\lambda$ is a tuneable hyper-parameter, which is set to 0.3 in this work. 
After training all ALR systems, we can obtain the transcripts transcribed from different systems.
Finally, we use ROVER, a post-recognition process that merges all transcripts into a single, minimum-cost word transition network (WTN) through iterative dynamic programming (DP) alignment.
The final WTN is searched by an automatic rescoring, or "voting," process to obtain the text sequence with the lowest error rate as our final text output.

\section{Experiment}
\subsection{Data Processing}
\textbf{Dataset}.
All ALR systems in this paper are built with the training and development dataset released by the ChatCLR Challenge Task 2.
The training, development, and evaluation sets include 110.95, 4.5, and 2.41 hours, respectively, all free talk video data, each recorded by 2 to 6 speakers in the TV living room.
The recording equipment is a 1080P resolution camera with a 120-degree wide-angle lens, placed 3 to 5 meters away from the speakers, allowing it to capture every speaker simultaneously.
For all recorded video data, the challenge official provides speakers' face and lip coordinates in almost every frame. \\ 
% In addition, all datasets provide ground-truth timestamps for the speakers.
% The training and development sets also offer text labels.
\textbf{Data Augmentation}.
During training, we randomly applied a variety of dynamic data augmentation combinations, including rotation, horizontal flipping, grayscale, and color jiggle, to each batch of video data. \\
\textbf{Speed Perturb}.
To enhance the robustness of our ALR systems, we apply triple speed perturbation at rates of 0.9, 1.0, and 1.1 to the training set videos using Moviepy~\footnote{https://pypi.org/project/moviepy}, an open-source Python library for video processing.

\subsection{Implementation Details}
\textbf{Model Configuration}.
All ALR systems are built based on the open-source toolkit Espnet~\cite{watanabe2018espnet}. 
To build diverse ALR systems, we use different scales of lip motion videos and visual encoders during training and inference. 
For the Branchformer encoder, we set the hidden size to 256, the number of encoder layers to 24, the dropout rate to 0.2, the cgmlp linear units to 2048, and the convolutional kernels to 31. 
The E-Branchformer includes 12 encoder layers, with the rest of the settings being the same as those for the Branchformer.
To compare fairly with the current mainstream Transformer and Conformer encoders, we adjust their parameters to ensure that different encoders have similar parameter sizes.
Specifically, we use a 24-layer Transformer and an 11-layer Conformer, both with a 2048 feed-forward dimension. 
% Other settings are consistent with the Branchformer.
All systems use the Enhanced ResNet3D visual front-end for feature extraction.
Four ResNet3D blocks have 3, 4, 6, and 3 basic blocks, respectively, with the feature dimensions of 32, 54, 128, and 256.
Six-layer Transformers, with 256 hidden sizes and a 0.2 dropout rate, serve as decoders in all ALR systems.\\
\textbf{Training Strategy}.
Each ALR system is trained for 50 epochs using the Adam optimizer with a learning rate of 1e-3, weight decay of 1e-6, and a linear warm-up strategy for the first 5 epochs.
% The learning rate increased linearly with the number of training steps. 
After the training, we average the top 20 epoch models with the lowest development loss as the decoding model on the development set and the initial model for the fine-tuning stage.
For fine-tuning, we use the same speakers' data in the training set as in the evaluation set and 90\% of the development set data as the training data, while the remaining 10\% as the fine-tuning development set. 
Set the learning rate to 5e-4 with the Adam optimizer, linearly warm-up for 5 epochs, and fine-tune for 20 epochs.
Finally, we average the top 10 epochs with the lowest loss on the fine-tuning development set as the decoding model for the evaluation set.\\
\textbf{Language Model and Inference Config}.
A 16-layer Transformer language model (LM) is built with text data from the training and development sets.
The embedding and attention dimensions are set to 128 and 512, respectively, and the model is trained for 30 epochs using the Adam optimizer with a learning rate of 0.001.
Finally, 10 models are averaged based on the development loss. 
During inference, the beam search decoding strategy is used with 64 beam-size.
The decoding CTC and LM weights are set to 0.35 and 0.1, respectively.

\begin{table}[t]
	\centering
	\caption{The CER(\%) results of ALR systems and ROVER outputs on development (Dev) and evaluation (Eval) sets.}
    \resizebox{\linewidth}{!}{
    \begin{tabular}{cccc|cc}
        \toprule
        System & Encoder & Scale & Param (M) & Dev $\downarrow$ & Eval $\downarrow$ \\ 
        \midrule
        S1 & Transformer & \multirow{4}*{0.6} & 53.3 & 83.09 & 87.61 \\
        S2 & Conformer & ~ & 50.8 & 83.61 & 87.14 \\
        S3 & E-Branchformer & ~ & 50.0 & 82.21 & 86.29 \\
        S4 & Branchformer & ~ & 52.6 & 81.97 & 85.87 \\
        \hline
        S5 & Branchformer & 0.8 & 52.6 & 80.35 & 84.69 \\
        \hline
        S6 & Transformer & \multirow{4}*{1.0} & 53.3 & 80.95 & 85.56  \\
        S7 & Conformer & ~ & 50.8 & 81.82 & 84.97 \\
        S8 & E-Branchformer & ~ & 50.0 & 79.02 & 83.44 \\
        S9 & Branchformer & ~ & 52.6 & 78.86 & 83.40 \\
        \hline
        S10 & Branchformer & 1.25 & 52.6 & 77.27 & 82.12 \\
        \hline
        S11 & Transformer & \multirow{4}*{1.5} & 53.3 & 79.13 & 84.97  \\
        S12 & Conformer & ~ & 50.8 & 79.88 & 83.51  \\
        S13 & E-Branchformer & ~ & 50.0 & 77.06 & 82.45  \\
        S14 & Branchformer & ~ & 52.6 & 76.21 & \textbf{81.56} \\
        \hline
        S15 & E-Branchformer & \multirow{2}*{1.75} & 50.0 & 76.01 & 82.61 \\
        S16 & Branchformer & ~ & 52.6 & \textbf{75.80} & 81.95 \\
        \midrule
        R1 & \multicolumn{3}{c|}{ROVER (S1 - S4)} & 78.78 & 83.15 \\
        R2 & \multicolumn{3}{c|}{ROVER (S6 - S9)} & 75.32 & 80.71 \\
        R3 & \multicolumn{3}{c|}{ROVER (S11 - S14)} & 73.49 & 79.58 \\
        R4 & \multicolumn{3}{c|}{ROVER (S4, S5, S9, S10, S14, S16)} & 73.45 & 79.58 \\
        R5 & \multicolumn{3}{c|}{ROVER ALL} & \textbf{72.02} & \textbf{78.17} \\
        \bottomrule
    \end{tabular}
    }
	\label{table-1}
\end{table}

\subsection{Main Results and Analysis}
\subsubsection{Which visual encoder performs best?}
To explore the influence of different visual encoders on the performance of the ALR system, we conduct comparative experiments at 3 video data scales, 0.6, 1.0, and 1.5. 
Table~\ref{table-1} shows detailed experimental results.
At the 0.6 and 1.0 scales, the ALR systems based on Branchformer (S4, S9) and E-Branchformer (S3, S8) encoders perform similarly, with Branchformers showing only 0.42\% and 0.04\% improvements in CER on the evaluation set at 0.6 and 1.0 scale, respectively.
However, compared to mainstream Transformers (S1, S6) and Conformers (S2, S7), Branchformer achieves noteworthy improvements of 1.74\% and 1.27\% at 0.6 scale, while 2.16\% and 1.57\% at 1.0 scale. 
At the 1.5 scale, Branchformer (S14) 's advantage is more pronounced, with reductions of 1.05\%, 1.95\%, and 3.41\% in CER on the evaluation set compared to E-Branchformer (S13), Conformer (S12), and Transformer (S11), respectively. 
In summary, the ALR system based on the Branchformer encoder shows the best performance, followed by the E-Branchformer, showing a big improvement over the Transformer and Conformer.

\subsubsection{Which data scale is most suitable?}
As analyzed in the previous section, the ALR system based on the Branchformer encoder performs the best. 
Therefore, we use the Branchformer as the visual encoder for our data scale comparison experiments, conducted at all 6 data scales: 0.6, 0.8, 1.0, 1.25, 1.5, and 1.75. 
The results are shown in Table~\ref{table-1}. 
It can be observed that within the range of 0.6 to 1.5 data scales, the larger the data scale used for model training and inference, the better the performance of the ALR system.
The system (S14) based on the Branchformer encoder built with 1.5-scale data shows a 4.31\% reduction in CER on the evaluation set compared to the one (S4) using 0.6-scale data.
However, as the data scale increases from 1.5 to 1.75, the ALR systems based on Branchformer (S16) and E-Branchformer (S15) encoders see their evaluation set CERs increase by 0.39\% and 0.16\%, respectively.
In summary, the ALR system built with lip motion video data at a scale of 1.5, under the same visual encoder, performs the best.

\subsubsection{How much gain does multi-system fusion bring?}
\label{multi-system-fusion-improvement}
To roundly show the benefits of multi-system fusion based on ROVER, we report not only the results of fusing all ALR systems (R5) but also the fusion of ALR systems built with different encoders (R1, R2, R3) and those using the same encoder but different scale data (R4), as shown in Table~\ref{table-1}.
First, we fuse ALR systems built with 0.6-scale video data based on four visual encoders. 
The fused result R1 shows a 2.72\% reduction in CER on the evaluation set compared to the best-performing ALR system (S4) at the 0.6 scale.
Similarly, the results fusing all ALR systems built with 1.0 (R2) and 1.5 scale data (R3) reduce CER by 2.69\% and 1.98\%, compared to the best-performing systems at the 1.0 scale (S9) and 1.5 scale (S14), respectively.
In addition, the result of fusing all ALR systems based on the Branchformer encoder but built with different scale data (R4) also shows a 1.98\% CER reduction compared to the best one (S14).
These results sufficiently demonstrate that significant improvements can be achieved through text fusion using ROVER, whether for ALR systems built with different visual encoders or data scales.
Finally, the result R5 that fused all ALR systems achieves a CER of 78.17\% on the evaluation set, which is a 3.39\% reduction compared to the best-performing system (S14), ranking second place in the ICME 2024 ChatCLR Challenge Task2.

\begin{table}[t]
	\centering
	\caption{The CER (\%) results of ablation experiments on development (Dev) and evaluation (Eval) sets.}
    \resizebox{\linewidth}{!}{
    \begin{tabular}{clc|cc}
        \toprule
        System & Method & Param (M) & Dev $\downarrow$ & Eval $\downarrow$ \\
        \midrule
        A0 & ResNet3D Baseline & 47.3 & 83.19 & 88.93 \\
        A1 & Enhanced ResNet3D & 52.6 & 81.23 & 87.45 \\
        A2 & + Data Augmentation & 52.6 & 76.27 & 83.84 \\
        A3 & ~~~+ Speed Perturb & 52.6 & 76.21 & 82.44 \\
        S14 & ~~~~~~+ Fine-tune & 52.6 & \textbf{76.21} & \textbf{81.56} \\
        \bottomrule
    \end{tabular}
    }
	\label{table-2}
\end{table}

\subsection{Ablation Study}
To prove the effectiveness of our proposed Enhanced ResNet3D visual front-end, data augmentation, speed perturbation, and fine-tuning on the performance of ALR systems, we conduct ablation experiments based on the Branchformer encoder using 1.5 scale video data. 
The ablation experiment results are shown in Table~\ref{table-2}. 
A0 is the ALR system based on the ResNet3D visual front-end and Branchformer proposed by Wang et al.~\cite{wang2024npu}, which we reproduce on the ChatCLR challenge dataset.
While A1 replaces the visual front-end with our proposed Enhanced ResNet3D, yielding a 1.48\% reduction in CER on the evaluation set. 
It is noteworthy that, comparing the results of systems A2 and A1, we find that incorporating the random data augmentation strategy during training yields significant benefits, with a 3.61\% reduction in CER on the evaluation set. 
Additionally, with speed perturbation on the training set, system A3 achieves a 1.4\% CER reduction compared to A2. 
Finally, after fine-tuning with training set data from the same speakers as the evaluation set and 90\% of the development set data, system S14 achieves the best single-system performance, further reducing the CER on the evaluation set by 0.88\% compared to system A3.

\section{Conclusion}
This paper summarizes our work in the target speaker lipreading task of the ICME2024 Chat-scenario Chinese Lip Reading (ChatCLR) challenge.
We propose a novel lip-reading approach that uses multi-scale lip motion video data and different visual encoders to build diverse automatic lip-reading (ALR) systems, and performs multi-system fusion on all output transcripts. 
For the ALR system, we propose an Enhanced ResNet3D visual front-end (VFE) module for extracting visual features, and introduce Branchformer and E-Branchformer as the visual encoder.
We report the results of all ALR systems and analyze three aspects: data, visual encoders, and multi-system fusion.
Ablation experiments are also conducted to show the effectiveness of the Enhanced ResNet3D VFE, data augmentation, speed perturbation, and fine-tuning.
Finally, we achieve a 78.17\% character error rate (CER) on the evaluation set with a 21.52\% reduction compared to the official baseline, ranking second place.

\bibliographystyle{IEEEbib}
\bibliography{main}

\begin{thebibliography}{10}

\bibitem{xiong2016achieving}
Wayne Xiong, Jasha Droppo, Xuedong Huang, Frank Seide, Mike Seltzer, Andreas Stolcke, Dong Yu, and Geoffrey Zweig,
\newblock ``{Achieving Human Parity in Conversational Speech Recognition},''
\newblock {\em arXiv preprint arXiv:1610.05256}, 2016.

\bibitem{vaswani2017attention}
Ashish Vaswani, Noam Shazeer, Niki Parmar, Jakob Uszkoreit, Llion Jones, Aidan~N Gomez, et~al.,
\newblock ``{Attention is All You Need},''
\newblock in {\em Proc. NIPS}. 2017, vol.~30, Curran Associates, Inc.

\bibitem{lee2020audio}
Yong-Hyeok Lee, Dong-Won Jang, Jae-Bin Kim, Rae-Hong Park, and Hyung-Min Park,
\newblock ``{Audio-visual Speech Recognition based on Dual Cross-modality Attentions with the Transformer Model},''
\newblock {\em Applied Sciences}, vol. 10, no. 20, pp. 7263, 2020.

\bibitem{serdyuk2021audio}
Dmitriy Serdyuk, Otavio Braga, and Olivier Siohan,
\newblock ``{Audio-visual Speech Recognition is Worth 32 X 32 X 8 Voxels},''
\newblock in {\em Proc. ASRU}. IEEE, 2021, pp. 796--802.

\bibitem{serdyuk2022transformer}
Dmitriy Serdyuk, Otavio Braga, and Olivier Siohan,
\newblock ``{Transformer-based Video Front-ends for Audio-Visual Speech Recognition for Single and Multi-person Video},''
\newblock {\em arXiv preprint arXiv:2201.10439}, 2022.

\bibitem{gulati2020conformer}
Anmol Gulati, James Qin, Chung-Cheng Chiu, Niki Parmar, Yu~Zhang, Jiahui Yu, Wei Han, Shibo Wang, et~al.,
\newblock ``{Conformer: Convolution-Augmented Transformer for Speech Recognition},''
\newblock in {\em Proc. Interspeech}. ISCA, 2020, pp. 5036--5040.

\bibitem{liu2022end}
Yixian Liu, Chuoya Lin, Mingchen Wang, Simin Liang, Zhuohui Chen, and Ling Chen,
\newblock ``End-to-end chinese lip-reading recognition based on multi-modal fusion,''
\newblock in {\em Proc. ICFTIC}. IEEE, 2022, pp. 794--801.

\bibitem{chang2024conformer}
Oscar Chang, Hank Liao, Dmitriy Serdyuk, Ankit Shahy, and Olivier Siohan,
\newblock ``{Conformer is All You Need for Visual Speech Recognition},''
\newblock in {\em Proc. ICASSP}. IEEE, 2024, pp. 10136--10140.

\bibitem{ma2021end}
Pingchuan Ma, Stavros Petridis, and Maja Pantic,
\newblock ``{End-to-End Audio-Visual Speech Recognition with Conformers},''
\newblock in {\em Proc. ICASSP}. IEEE, 2021, pp. 7613--7617.

\bibitem{ma2023auto}
Pingchuan Ma, Alexandros Haliassos, Adriana Fernandez-Lopez, Honglie Chen, et~al.,
\newblock ``{Auto-AVSR: Audio-Visual Speech Recognition with Automatic Labels},''
\newblock in {\em Proc. ICASSP}. IEEE, 2023, pp. 1--5.

\bibitem{chen2022first}
Hang Chen, Hengshun Zhou, Jun Du, Chin-Hui Lee, Jingdong Chen, Shinji Watanabe, Sabato~Marco Siniscalchi, Odette Scharenborg, et~al.,
\newblock ``{The first Multimodal Information based Speech Processing (MISP) Challenge: Data, Tasks, Baselines and Results},''
\newblock in {\em Proc. ICASSP}. IEEE, 2022, pp. 9266--9270.

\bibitem{wang2023multimodal}
Zhe Wang, Shilong Wu, Hang Chen, Mao-Kui He, Jun Du, Chin-Hui Lee, Jingdong Chen, et~al.,
\newblock ``{The Multimodal Information based Speech Processing (MISP) 2022 Challenge: Audio-Visual Diarization and Recognition},''
\newblock in {\em Proc. ICASSP}. IEEE, 2023, pp. 1--5.

\bibitem{wu2024multimodal}
Shilong Wu, Chenxi Wang, Hang Chen, Yusheng Dai, Chenyue Zhang, Ruoyu Wang, Hongbo Lan, Jun Du, Chin-Hui Lee, Jingdong Chen, et~al.,
\newblock ``{The Multimodal Information based Speech Processing (MISP) 2023 Challenge: Audio-Visual Target Speaker Extraction},''
\newblock in {\em Proc. ICASSP}. IEEE, 2024, pp. 8351--8355.

\bibitem{chen2022audio}
Hang Chen, Jun Du, Yusheng Dai, Chin~Hui Lee, Sabato~Marco Siniscalchi, Shinji Watanabe, et~al.,
\newblock ``{Audio-Visual Speech Recognition in MISP2021 Challenge: Dataset Release and Deep Analysis},''
\newblock in {\em Proc. Interspeech}. ISCA, 2022, pp. 1766--1770.

\bibitem{chen2023cn}
Chen Chen, Dong Wang, and Thomas~Fang Zheng,
\newblock ``{CN-CVS: A Mandarin Audio-Visual Dataset for Large Vocabulary Continuous Visual to Speech Synthesis},''
\newblock in {\em Proc. ICASSP}. IEEE, 2023, pp. 1--5.

\bibitem{wang2024mlca}
He~Wang, Pengcheng Guo, Pan Zhou, and Lei Xie,
\newblock ``{MLCA-AVSR: Multi-Layer Cross Attention Fusion Based Audio-Visual Speech Recognition},''
\newblock in {\em Proc. ICASSP}, 2024, pp. 8150--8154.

\bibitem{wang2024npu}
He~Wang, Pengcheng Guo, Wei Chen, Pan Zhou, and Lei Xie,
\newblock ``{The NPU-ASLP-LiAuto System Description for Visual Speech Recognition in CNVSRC 2023},''
\newblock {\em arXiv preprint arXiv:2401.06788}, 2024.

\bibitem{peng2022branchformer}
Yifan Peng, Siddharth Dalmia, Ian Lane, and Shinji Watanabe,
\newblock ``{Branchformer: Parallel MLP-Attention Architectures to Capture Local and Global Context for Speech Recognition and Understanding},''
\newblock in {\em Proc. ICML}. PMLR, 2022, pp. 17627--17643.

\bibitem{kim2023branchformer}
Kwangyoun Kim, Felix Wu, Yifan Peng, Jing Pan, Prashant Sridhar, et~al.,
\newblock ``{E-branchformer: Branchformer with Enhanced Merging for Speech Recognition},''
\newblock in {\em Proc. SLT}. IEEE, 2023, pp. 84--91.

\bibitem{fiscus1997post}
Jonathan~G Fiscus,
\newblock ``{A Post-processing System to Yield Reduced Word Error Rates: Recognizer output voting error reduction (ROVER)},''
\newblock in {\em Proc. ASRU}. IEEE, 1997, pp. 347--354.

\bibitem{he2016deep}
Kaiming He, Xiangyu Zhang, Shaoqing Ren, and Jian Sun,
\newblock ``{Deep Residual Learning for Image Recognition},''
\newblock in {\em Proc. CVPR}. IEEE/CVF, 2016, pp. 770--778.

\bibitem{zhang2020can}
Yuanhang Zhang, Shuang Yang, Jingyun Xiao, Shiguang Shan, and Xilin Chen,
\newblock ``{Can we read speech beyond the lips? Rethinking ROI Selection for Deep Visual Speech Recognition},''
\newblock in {\em 2020 15th IEEE International Conference on Automatic Face and Gesture Recognition (FG 2020)}. IEEE, 2020, pp. 356--363.

\bibitem{kim2017joint}
Suyoun Kim, Takaaki Hori, and Shinji Watanabe,
\newblock ``{Joint CTC-Attention based End-to-End Speech Recognition using Multi-task Learning},''
\newblock in {\em Proc. ICASSP}. IEEE, 2017, pp. 4835--4839.

\bibitem{watanabe2018espnet}
Shinji WWatanabe, Takaaki Hori, Shigeki Karita, Tomoki Hayashi, Jiro Nishitoba, Yuya Unno, Nelson Enrique~Yalta Soplin, Jahn Heymann, et~al.,
\newblock ``{Espnet: End-to-End Speech Processing Toolkit},''
\newblock in {\em Proc. Interspeech}. ISCA, 2018, pp. 2207--2211.

\end{thebibliography}

\end{document}